\pdfoutput=1

\documentclass[11pt]{article}

\usepackage[]{ACL2023} 
\usepackage{times}
\usepackage{latexsym}

\usepackage[T1]{fontenc}

\usepackage[utf8]{inputenc}

\usepackage{microtype}

\usepackage{inconsolata}

\usepackage{adjustbox}
\usepackage{tabularx}
\usepackage{makecell}
\usepackage{multirow}
\usepackage{enumitem}
\usepackage{caption}
\usepackage{bbding}
\usepackage{color}
\usepackage{booktabs}
\usepackage{xcolor}

\title{Check-COVID: Fact-Checking COVID-19 \\ News Claims with Scientific Evidence}

\author{Gengyu Wang\textsuperscript{1} ~~ Kate Harwood \textsuperscript{1} ~~ Lawrence Chillrud \textsuperscript{2} \\ \bf{Amith Ananthram \textsuperscript{1} ~~ Melanie Subbiah\textsuperscript{1}  ~~ Kathleen McKeown\textsuperscript{1}} \\
\textsuperscript{1}Columbia University ~~ 
\textsuperscript{2}Northwestern University \\ 
\texttt{\{gengyu.wang, k.r.harwood\}@columbia.edu, chili@u.northwestern.edu}\\  \texttt{\{amith.ananthram, m.subbiah\}@columbia.edu, kathy@cs.columbia.edu}}
 
\begin{document}
\maketitle
\begin{abstract}
We present a new fact-checking benchmark, Check-COVID, that requires systems to verify claims about COVID-19 from news using evidence from scientific articles. This approach to fact-checking is particularly challenging as it requires checking internet text written in everyday language against evidence from journal articles written in formal academic language. Check-COVID contains $1,504$ expert-annotated news claims about the coronavirus paired with sentence-level evidence from scientific journal articles and veracity labels. It includes both \textit{extracted} (journalist-written) and \textit{composed} (annotator-written) claims. Experiments using both a fact-checking specific system and GPT-3.5, which respectively achieve F1 scores of 76.99 and 69.90 on this task, reveal the difficulty of automatically fact-checking both claim types and the importance of in-domain data for good performance. Our data and models are released publicly at \url{https://github.com/posuer/Check-COVID}.

\end{abstract}

\section{Introduction}
\label{sec_intro}

\begin{figure}[ht!]
\includegraphics[scale=0.72]{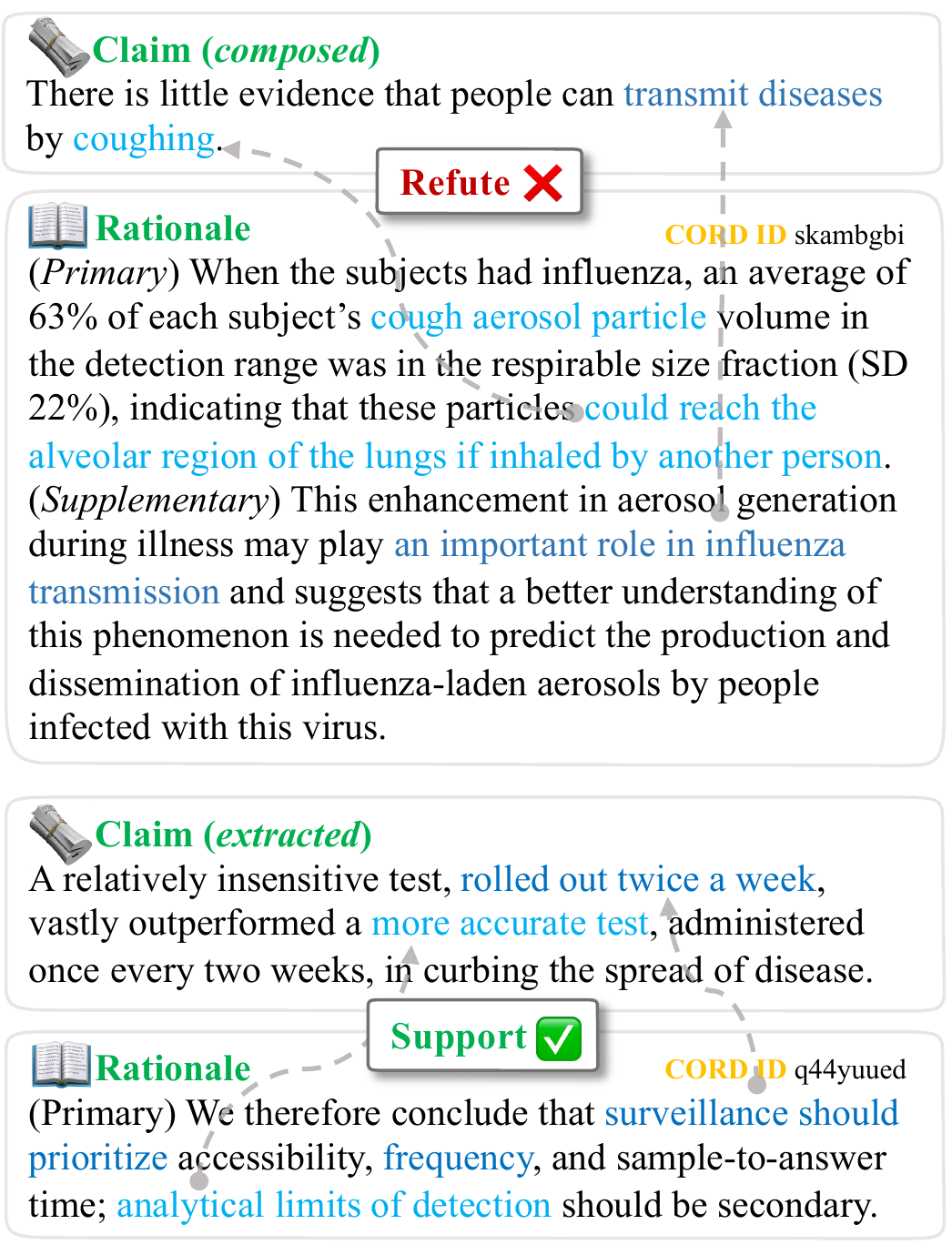}
\centering
\caption{Check-COVID examples. \textit{Composed} claims are annotator-written based on assertions in news articles, whereas \textit{extracted} claims are copied verbatim.
}
\label{table_covid}
\end{figure}
Throughout the COVID-19 pandemic, misinformation on the internet has proven to be exceptionally dangerous, undercutting containment efforts by public health officials around the world \cite{mheidly2020leveraging}. In future pandemics, the ability to automatically detect and debunk such misinformation has the potential to save many lives. However, an automated system 
cannot rely solely on surface forms or linguistic features to identify misinformation \cite{perez2017automatic, rashkin2017truth}. Such a system must 
be capable of checking claims written in 
everyday
language against jargon-laden evidence from an evolving set of scientific articles.

In this work, we formalize this challenge as a new fact-checking benchmark that requires verifying claims about COVID-19 from news using evidence from scientific articles.  This task is particularly challenging as it requires grounding everyday vernacular in formal academic language.  Consider the logical inferences required to correctly label the first example presented in Figure \ref{table_covid} as refuted by the evidence. An automated system must recognize that \textit{coughing} produces \textit{aerosol particles} that may \textit{reach the alveolar region of the lungs of another person}, which implies infection of that person. A successful model must be able to align 
such everyday
language with scientific terminology while leveraging commonsense knowledge.

To facilitate research on this task, we introduce Check-COVID, a fact-checking dataset of $1,504$ claims about COVID-19, each of which is paired with sentence-level scientific evidence for its veracity label. These claims are drawn from news articles, and the evidence is selected by human annotators from a large COVID-19 biomedical literature corpus, CORD-19 \cite{wang-2020-cord}. The dataset includes $322$ \textit{extracted} claims, with wording drawn directly from the article, and $1,182$ \textit{composed} claims, re-worded by trained annotators.
Other fact-checking datasets have focused solely on one type of claim or the other. However, both types are important since people often include direct quotes or their own re-wording of claims when sharing articles online. Thus our dataset allows us to explore how models handle both these real-world claim types within the same domain. 
To build this dataset, we 
adapt SciFact's annotation method \cite{wadden2020fact} to news, leveraging naturally occurring citations in articles. Such citations usually contain claims with links to supporting journal articles. We first collect these claims and the abstracts of the referenced scientific articles and then manually identify the sentences in each abstract that provide claim-specific evidence.

To establish a baseline score, we adapt the fact-checking system presented in \citet{deyoung2019eraser} and \citet{wadden2020fact}, which uses an abstract retriever, rationale selector, and label predictor.
To augment our training data, we experiment with existing fact-checking datasets that use a similar task formulation but contain claims from Wikipedia articles (FEVER \cite{thorne2018fever}), scientific journal articles (SciFact \cite{wadden2020fact}), or Reddit (COVID-Fact \cite{saakyan2021covid}) in addition to our corpus, Check-COVID. We train the models on different combinations of these four datasets, compare their performance, and select the best performing models to generate the baseline score.  Experimental results make clear the difficulty of adapting existing corpora to this new benchmark - their inclusion yields only small improvements in performance. Moreover, an evaluation of GPT-3.5 \cite{brown2020language} reveals limitations of in-context learning, especially in providing human-aligned evidence for its veracity labels.

Our contributions are: 1) we introduce a novel dataset (Check-COVID) as a benchmark for the challenging problem of fact-checking COVID-19 claims from news against evidence from scientific journal articles; 2) we evaluate how well models can adapt to both \textit{extracted} and \textit{composed} claims within the same domain; and 3) we present a strong baseline for Check-COVID to facilitate future work.
Our data and models will be released publicly under the MIT License at \url{https://github.com/posuer/Check-COVID}.

\section{Related Work}

There has been tremendous interest in developing fact-checking benchmarks for COVID-19, however many of these datasets only contain claims (\citet{shah2022concord}) or claims with veracity labels and no evidence (\citet{li2022covid, elhadad2020covid, shahi2020fakecovid, cui2020coaid, vijjali2020two, alam2020fighting}).  Others primarily source claims from social media (\citet{saakyan2021covid}; \citet{mohr2022covert}; \citet{sundriyal2022document}) or scientific texts (\citet{wadden2020fact}). 
Some datasets, like ours, source claims from colloquial news sources (or other online sources) (\citet{lee2020misinformation}; \citet{lee2021towards}; \citet{sarrouti2021evidence}) but of these \citet{lee2020misinformation} and \citet{lee2021towards} have just a few hundred examples, a challenge for use in training deep learning systems. 

\citet{sarrouti2021evidence} is the most similar to our dataset, however, our evidence comes from citations in the news articles the claims are drawn from, whereas in \citet{sarrouti2021evidence} the evidence is retrieved using the extracted claims as search queries.  Thus, our claim-evidence pairings are closer to those the average reader encounters online.

Additionally we include a third veracity label (\texttt{NOTENOUGHINFO}) which some of the above datasets - and many general fact-checking corpora - do not.  In the wild, a system will not always be able to fact-check a claim.  Thus, modeling these cases is critical to real-world performance.

\section{Check-COVID Dataset}

We introduce Check-COVID, a dataset of $1,504$ claims drawn from news paired with sentence-level evidence from scientific journal articles and accompanying veracity labels: $\{\texttt{SUPPORT,}$ $\texttt{REFUTE,}$ $\texttt{NOTENOUGHINFO}\}$. The number of examples across the three labels is balanced ($505, 504, 495$). The claims are categorized into \textit{composed} or \textit{extracted} based on whether they are written by our annotators or drawn from news articles.
We also provide a corpus containing the abstracts of the journal articles from which the evidence is drawn, a subset of the CORD-19 corpus. We randomly split the dataset into three balanced subsets with no overlapping abstracts: train ($70\%$), dev ($15\%$) and test ($15\%$).

\subsection{Data Source}
Citances \cite{nakov2004citances} in news are spans of text? that contain assertions about findings from scientific journals with accompanying citations to the articles where the supporting evidence can be found. We build a crawler to automatically detect citations in news and collect sentences surrounding the citance, the abstract of the cited journal article, and the articles' URLs. To ensure the cited information is scientific, the crawler ignores citances which do not cite journal articles in CORD-19,  a trustworthy corpus of scientific papers on coronavirus research.
We restrict our corpus to a set of well-regarded news websites\footnote{Including The New York Times, The Washington Post, and BBC. The full list is in the Appendix \ref{sec:News_Citance_Sources}.} to increase the likelihood that claims are paired with relevant scientific evidence (fake news, in contrast, regularly contains deceptive citances). Our manual annotation and claim negation process is then a second check to ensure that cited evidence is appropriate.

\subsection{Claims in Check-COVID}

In Check-COVID, a claim is an atomic factual statement describing 
one aspect of a scientific entity or process related to COVID-19 (such that it can be fact-checked against primary research).  For example, ``\textit{Cloth masks offer significantly less protection against infection than medical masks.}" Opinions or facts that do not require scientific proof are not considered valid claims (e.g., 
``\textit{The government published a policy requiring masks in public spaces}'').

\textbf{Composed and Extracted} Check-COVID contains both \textit{composed}  and \textit{extracted} claims. 
To generate \textit{composed}  claims,
we present annotators with news paragraphs that contain a citance. Annotators are asked to write a claim based on the information in the paragraph. We require that \textit{composed} claims be understandable without extra context. 
For \textit{extracted} claims, we detect claims from citance-containing news paragraphs using a claim detection model \cite{barron2019proppy} trained on ClaimBuster \cite{arslan2020benchmark}, then we present them to annotators to decide whether they satisfy our definition of a claim. 
In Figure \ref{table_covid}, we present two claims, one \textit{composed} and one \textit{extracted}. \textit{Composed}  claims are usually shorter, simpler, and more similar to the kind of claims that the general public might write online about an article or submit to a fact-checking system.  
In contrast, \textit{extracted} claims are retrieved directly from news articles, often use more complicated wording, and are necessary to train systems that fact-check news directly. Our experiments demonstrate that models trained on one category cannot simply be adapted to the other. In Check-COVID, the ratio between \textit{composed}  claims and \textit{extracted} claims is $3.67:1$. We explore differences between them in \S \textbf{Dataset Statistics} and Table \ref{table:stat}.

\textbf{Claim Negation} To create examples that are refuted by the cited abstracts, we manually negate the supported claims.
Negation procedures can introduce bias in a dataset which can allow a model to ``cheat'' on a task. For example, a model can learn to associate the word ``not" with the \texttt{REFUTE} label \cite{schuster2019towards}. To mitigate these effects, we request that annotators avoid using negators like “does not”, “cannot”, “no”, etc. Instead, we ask them to negate claims by changing them more fundamentally, for example, by  changing \textit{Cloth masks offer much less protection against infection than medical masks} to \textit{Cloth masks are just as effective at preventing infection as medical masks}. As negations involve minimal changes to the original claim style, negated claims retain the type designation (\textit{composed} or \textit{extracted}) of their originals.

\subsection{Annotation Procedure}

On a web-based annotation interface (see Appendix for screenshots), annotators are shown a news paragraph or a selection of automatically detected claims together with one of the paragraph’s cited abstracts. 
Annotators start by either selecting a valid claim from those provided or writing a \textit{composed} claim from the full news paragraph. In either case, they then write a second claim that is the negation of the first.  For each claim, annotators identify whether the claim is ``\texttt{SUPPORTED}" or ``\texttt{REFUTED}" by the abstract and select rationales from the abstract as justification. A rationale is a minimal collection of sentences sufficient to justify a label of \texttt{SUPPORT} or \texttt{REFUTE} in relation to a claim.

In fact-checking, often there is not adequate evidence to confidently certify or debunk a claim, so we also introduce a third label, \texttt{NOTENOUGHINFO}. To create a \texttt{NOTENOUGHINFO} example, we randomly sample from the \textit{composed} claims and pair the sampled claim with a sentence from its corresponding abstract that was not chosen by an annotator as evidence for that claim. By choosing non-evidence sentences from the cited abstract, we select sentences that exhibit both lexical and topic overlap with the claim without providing evidence for it, resulting in difficult \texttt{NOTENOUGHINFO} examples. \subsection{Quality Control}

We employ four graduate students with a background in NLP and four graduate students studying life sciences as annotators through a mailing list of the education institution. We paid them the minimum hourly rate required by our locality and obtained signed informed consent forms that explain how the collected data would be used. We obtained approval (ethics review) for this study from the Institutional Review Board of the authors' institution. As our data is collected or derived from publicly available sources, it is not subject to any anonymization practices. We removed any identifying information from the annotations.
All annotators watch a video, read an instruction guide and produce practice annotations to become familiar with the task. We include the instructions in the Appendix \ref{sec:annotation_interface} and we make the video available online\footnote{\url{https://youtu.be/mEOcouML9oA}}.
As annotators need to write their own versions of claims, it is difficult to calculate an agreement score for their annotations. Therefore, to control quality, each annotator is assigned to an NLP graduate student who reviews each annotation, provides revision suggestions, and monitors revisions until the annotation meets the requirements described above. 
Approved annotators then review each other's submissions.  At the same time, expert annotators continue monitoring new annotations for quality and provide feedback when necessary. As a final check, all submitted claims are proofread by the authors. We believe such adjudication, though laborious, is of renewed importance in our field, providing high quality benchmarks that accurately measure model performance \cite{pustejovsky2012natural}.

\subsection{Dataset Statistics}

\iftrue
\begin{figure}[] 
\includegraphics[scale=0.12]{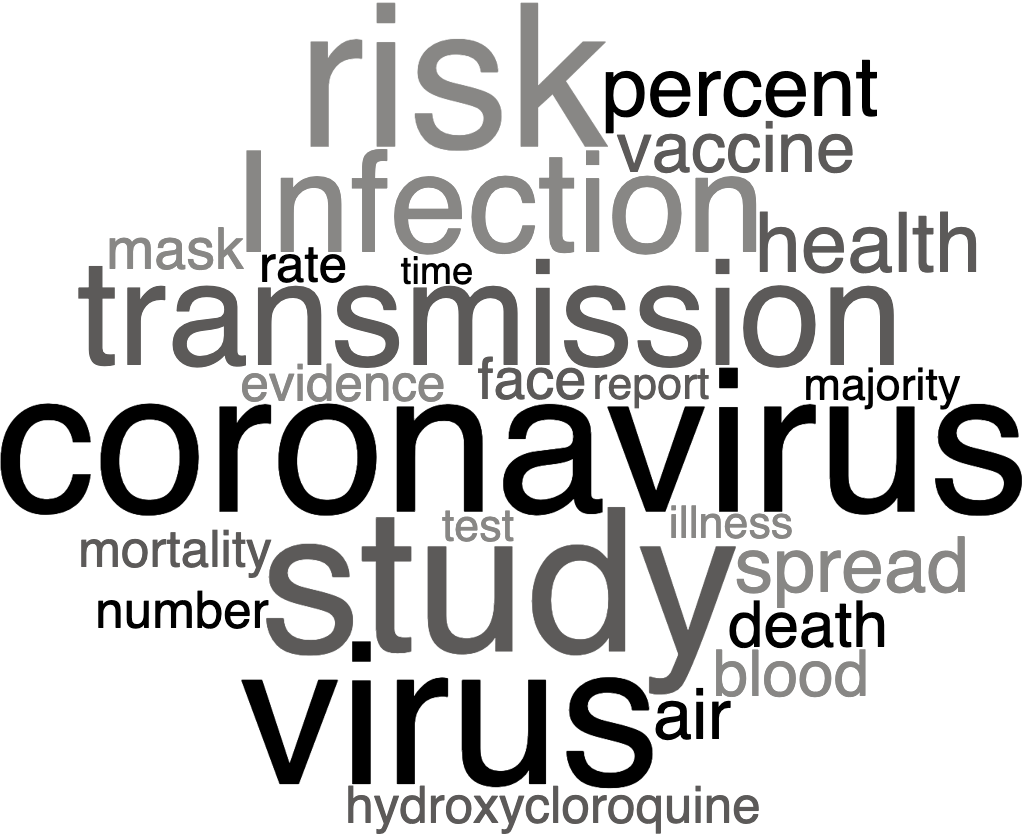}
\centering
\caption{Common words in Check-COVID claims.}
\vspace{-5mm}
\label{figure:word_cloud}
\end{figure}
\fi
\label{sec:stat}
Check-COVID covers a diverse set of topics as demonstrated by the variety of keywords in claims shown in Figure \ref{figure:word_cloud}. Table \ref{table:stat} presents summary statistics for our corpus. We can see that \textit{extracted} claims are twice as long as \textit{composed} claims, suggesting that \textit{composed} claims use simpler wording. This linguistic diversity, along with the differences in system performance in Section \ref{sec_experiments}, is further evidence for the importance of including both claim types as part of a fact-checking benchmark.

Since we annotate \texttt{REFUTE} claims by manually negating \texttt{SUPPORT} claims, the ``not/no'' in \texttt{REFUTE} claims are either from their source \texttt{SUPPORT} claims or introduced by the annotation process.
While we request that annotators avoid using ``not/no'' during negation, the percentage of claims containing ``not/no" shows that more \texttt{REFUTE} claims contain ``not/no'' than \texttt{SUPPORT} claims. However, our results for label prediction in Table \ref{tabel_covid-lab} show that this correlation does not result in a clear pattern of \texttt{REFUTE} claims being easier to check than \texttt{SUPPORT} claims.

\begin{table}[]
\centering
\begin{adjustbox}{max width=0.47\textwidth}
\begin{tabular}{l|ccc}
\toprule
  & \texttt{REFUTE} & \texttt{NOTENINFO} & \texttt{SUPPORT}       \\
\midrule
\textbf{Composed}  \\
Number of claims & 419 & 392 & 371 \\
Avg. \# words per claim & 16.54 & 16.66 &  16.72  \\
\% claims containing \textit{not/no} & 0.32 & 0.24 & 0.10 \\
Avg. \# sentences per rationale & 1.45 & 1.00 & 1.46 \\
Avg. \# words per evidence sentences & 33.32 & 28.64 & 33.20 \\
\midrule
\textbf{Extracted} \\
Number of claims & 85 & 103 & 134 \\
Avg. \# words per claim &  29.95  & 29.35 & 29.03 \\
\% claims containing \textit{not/no} & 0.35 & 0.19 & 0.13 \\
Avg. \# sentences per rationale & 1.54 & 1.0 & 1.57 \\
Avg. \# words per evidence sentence & 33.04 & 26.94 & 32.98 \\
\bottomrule
\end{tabular}
\end{adjustbox}
\caption{Check-COVID summary statistics. 
}
\vspace{-5mm}
\label{table:stat}
\end{table}

\section{Check-COVID Task and Challenges}

Given a claim about COVID-19, either written by an annotator or drawn from a news article,
the task is to retrieve related scientific abstracts from CORD-19, choose the most relevant evidence sentences from those abstracts and decide whether that evidence supports, refutes, or does not provide enough information about the claim. This formulation matches the SciFact task format. The claims in our corpus pose many unique challenges:

\textbf{Medical Knowledge and Terminology} 
Many examples in our corpus require knowledge of medicine and its terminology 
to perform successful fact-checking. To fact-check the first example in Figure \ref{table_covid}, a model must recognize that the \textit{alveolar region} is \textit{where blood exchanges oxygen and carbon dioxide}\footnote{from NCI: \url{https://www.cancer.gov/publications/dictionaries/cancer-terms/def/alveoli}}. This implies that 
here, the virus can enter someone's bloodstream and infect them.

\textbf{Temporal Reasoning}
The examples in Check-COVID also require understanding and comparing 
temporal durations, frequencies and ordering. To fact-check the second claim in Figure \ref{table_covid}; a model needs to understand that testing \textit{twice a week} is more frequent than \textit{once every two weeks}.

\textbf{Numeric Values}
Most examples with numeric values require numeric comparison and number-word translation. 
For example, given the claim \textit{communities of color were disproportionately affected by COVID-19} and its evidence \textit{cumulative incidence was... higher among Hispanic/Latino (29.2\%)... than non-Hispanic white adults (8.1\%, p\textless.0001)}, a model must compare \textit{29.2\%} and \textit{8.1\%} and deem this difference \textit{disproportionate}. 

We closely scrutinize the prevalence and impact of these challenges. With regard to medical terminology and knowledge, all examples in our dataset pose this challenge, as our task requires fact-checkers to understand and extract evidence from medical journal articles. Regarding temporal reasoning and numeric reasoning, there are 147 and 93 claims across a total of 450 dev and test set examples respectively that require such reasoning to be fact-checked. Although we will delve deeper into the fact-checking system and related experiments in forthcoming sections, our analysis indicates a decline in the system's performance on this specific subset. Notably, the macro F1 of the composed test set processed through the Vespa pipeline fell from 63.34 to 29.80 for the temporal subset, and to 33.87 for the numeric subset.

\section{Baseline System}

Our fact-checking pipeline is adapted from the systems presented in \citet{deyoung2019eraser}, \citet{wadden2020fact}, and \citet{wang2021climate}, whose use is consistent with their intended use. It consists of three modules, trained in a supervised fashion on different combinations of existing fact-checking corpora and Check-COVID.
We describe the individual components of this baseline system below.

\textbf{Abstract Retrieval} 
We query for COVID-19 scientific literature through a CORD-19 search engine, \textit{Vespa}\footnote{https://cord19.vespa.ai/}. Given a claim, \textit{Vespa} retrieves relevant abstracts from CORD-19 and ranks them with the BM25 scoring function \cite{robertson1995okapi}. By integrating \textit{Vespa} instead of using a retrieval model with a fixed corpus, we enable the pipeline to retrieve evidence from a corpus of COVID-19 scientific articles that could be continually updated. The three abstracts with the highest BM25 scores are passed to the next step of \textit{Rationale Selection}.

\textbf{Rationale Selection}
We fine-tune a RoBERTa \cite{liu2019roberta} model with Transformers \cite{wolf-etal-2020-transformers-1} to predict whether each sentence $e_i$ from the selected abstracts is relevant to the given claim $c$, where $i$ refers to the index within the abstract. We encode each sequence [$c$ $SEP$ $e_i$] and then feed the final $[CLS]$ representation to a linear classification layer to predict a binary relevance label. Sentences that score over $0.5$ are passed as a (possibly empty) set  to \textit{Label Prediction}.

\textbf{Label Prediction}
We fine-tune another RoBERTa model to label each selected rationale $e_i$ as $\texttt{SUPPORT}$, $\texttt{REFUTE}$, or $ \texttt{NOTENOUGHINFO}$.  
Given a claim $c$ and evidence $e_i$, we encode the sequence [$c$ $SEP$ $e_i$] with RoBERTa and then feed the final $[CLS]$ representation to a linear classification layer. We optimize a cross-entropy loss over our $3$ classes. For claims with a multi-sentence rationale, $e_i$ is the concatenation of the sentences.

\begin{table*}[t]
\centering
\begin{adjustbox}{max width=0.9\textwidth}
\begin{tabular}{lcccccc|cc}
\toprule
\multirow{3}{*}{\textbf{Rationale Selection}}  & \multicolumn{6}{c}{\textbf{Sentence Level}} &  \multicolumn{2}{c}{\textbf{Rationale Level}}\\
 & \multicolumn{3}{c}{\textbf{Standard}}                                                                        & \multicolumn{3}{c}{\textbf{Global Recall-Focused}}                                  & \textbf{Strict}  & \textbf{Intersection}                                   \\
          &   Precision & Recall & F1 & Precision & Recall & F1 & Accuracy & Accuracy \\

\midrule
\midrule
\multicolumn{9}{c}{Evaluated on \textbf{Composed}}\\
\midrule
\multicolumn{3}{l}{  \textit{* Train on composed}}           \\
Check-COVID                   & 53.20 & 34.72 & 39.95 (6.31) & 44.80 & 28.24 & 32.91 (2.34) & 19.81 (1.89) & 44.65 (16.05) \\
FEVER + Check-COVID           & 47.56 & 53.47 & 50.25 (1.03) & 34.82 & 39.12 & \textbf{36.78}\dag (2.26) & 22.96 (3.93) & \textbf{69.81} (2.50) \\
SciFact + Check-COVID         & 52.67 & 45.14 & 48.26 (1.66) & 38.28 & 32.87 & 35.12 (2.18) & 24.53 (5.25) & 60.06 (5.76)\\
FEVER + SciFact + Check-COVID & 54.38 & 49.07 & \textbf{51.23} (1.83) & 38.90 & 34.95 & 36.56 (1.14) & \textbf{\underline{29.25}} (2.50) & 64.78 (7.57) \\
\midrule
\multicolumn{3}{l}{ \textit{* Train on composed  \textbf{and} extracted together}}  \\
Check-COVID                   & 47.75 & 47.69 & 47.27 (0.96)& 38.69 & 38.66 & \textbf{\underline{38.31}}\dag(0.54) & 23.27 (6.82) & 62.26 (5.74) \\
FEVER + Check-COVID           & 50.87 & 53.01 & 51.70 (1.06) & 37.14 & 38.89 & 37.83 (2.87) & 26.42 (1.63) & 68.87 (5.74) \\
SciFact + Check-COVID         & 53.80 & 50.46 & 52.05 (1.74) & 38.28 & 35.88 & 37.02 (2.70) & \textbf{26.73} (1.44) & 66.35 (1.44)\\
FEVER + SciFact + Check-COVID & 47.76 & 57.87 & \textbf{\underline{52.18}} (0.40) & 33.99 & 41.20 & 37.14 (0.88) & 24.84 (3.31) & \textbf{\underline{75.16}} (5.20) \\
 \\
 \midrule
GPT-3.5 (text-davinci-003) & 51.35 & 39.58  & 44.71 & 42.34 & 32.64 & 36.86 & 16.98 & 49.06
 \\
\midrule
\midrule
\multicolumn{9}{c}{Evaluated on \textbf{Extracted}}  \\
\midrule
\multicolumn{3}{l}{\textit{* Train on extracted}}            \\

Check-COVID                   & 72.49  & 46.26  & 56.39 (1.37)  & 36.05  & 23.13  & 28.13 (4.00)  & 25.25 (3.78) & 62.63 (1.43) \\
FEVER + Check-COVID           & 73.16 & 59.18 & \textbf{\underline{65.40}} (0.90) & 49.69 & 40.14 & \textbf{\underline{44.39}}\dag(3.05) & \textbf{\underline{41.41}} (1.75) & \textbf{79.80} (1.75) \\
SciFact + Check-COVID         & 70.30 & 50.34 & 58.62 (6.05) & 39.93 & 28.57 & 33.28 (4.06) & 31.31 (3.50) & 67.68 (8.75) \\
FEVER + SciFact + Check-COVID & 65.56 & 53.06 & 58.21 (1.76) & 46.16 & 37.41 & 41.01 (2.47) & 31.31 (3.50) & 73.74 (4.63) \\
\midrule
\multicolumn{3}{l}{\textit{* Train on composed  \textbf{and} extracted together}}  \\
Check-COVID                   & 56.44 & 57.82 & 56.52 (1.08) & 36.98 & 38.78 & 37.46 (5.35) & 24.24 (3.03) & 71.72 (4.63)\\
FEVER + Check-COVID           & 69.43 & 55.10 & \textbf{61.38} (2.16) & 39.28 & 31.29 & 34.80 (2.29) & \textbf{36.36} (3.03) & 75.76 (3.03) \\
SciFact + Check-COVID         & 57.31 & 52.38 & 54.42 (3.97) & 31.98 & 29.25 & 30.38 (2.03) & 32.32 (3.50) & 71.72 (12.25) \\
FEVER + SciFact + Check-COVID & 58.45 & 63.95 & 60.90 (1.27) & 38.92 & 42.86 & \textbf{40.67}\dag (2.24) & 30.30 (5.25) & \textbf{\underline{80.81}} (1.75) \\
\midrule
GPT-3.5 (text-davinci-003) & 58.33 & 42.86  & 49.41  & 36.11 & 26.53  & 30.59 & 18.18 & 57.58\\
\bottomrule
\end{tabular}
\end{adjustbox}
\caption{Results (avg. of $3$ seeds, std in parentheses) for \textit{Rationale Selection} with different configurations of \textit{composed} and \textit{extracted} claims from Check-COVID and with or without FEVER and SciFact for training data. Metric details in Section \ref{sec:rateval}.}\vspace{-3mm}
\label{tabel:covid-rat}
\end{table*}

\begin{table*}[]
\centering
\begin{adjustbox}{max width=1\textwidth}
 \begin{tabular}{lcccc|cccc}
\toprule
\textbf{Label Prediction} & \multicolumn{4}{c}{\textbf{Composed}}  & \multicolumn{4}{c}{\textbf{Extracted}} \\
& \texttt{REFUTE} &	\texttt{NOTENINFO} &	\texttt{SUPPORT}	 & Macro F1 (std) &\texttt{REFUTE} &	\texttt{NOTENINFO} &	\texttt{SUPPORT}	 & Macro F1 \\
                              \midrule
\multicolumn{3}{l}{ \textit{* Train \& evaluate on composed \textbf{or} extracted separately}}            \\
Check-COVID                   & 85.86 &	78.91 &	81.30 & 82.03 (1.47) & 45.01 &	68.25 &	70.94 &   61.40 (14.41)\\
FEVER + Check-COVID           & 85.03 &	80.24 &	84.85 & 83.37 (0.93) & 84.30 &	87.03 &	84.25 & \textbf{\underline{85.19}}\dag (2.11) \\
SciFact + Check-COVID         & 82.19 &	79.00 &	81.26 & 80.82 (1.29) & 77.33 &	80.49 &	82.17 & 80.00 (2.24 \\
FEVER + SciFact + Check-COVID &   85.00 &	82.05 &	84.21 & \textbf{\underline{83.76}}\dag  (0.63) & 82.76 &	78.44 &	82.77 & 81.33 (1.29)  \\
COVID-Fact + Check-COVID     & 84.58 &	78.49 &	85.73 & 82.93 (0.99) & 38.75 &	34.64 &	57.73 & 43.70 (25.17)\\
\midrule
\multicolumn{3}{l}{ \textit{* Train on composed \textbf{and} extracted together.}}           \\
Check-COVID                   & 83.78 & 76.91 &	79.74 & 80.14 (2.31) & 74.59 &	89.92 &	84.63 & \textbf{83.05}\dag (1.80) \\
FEVER + Check-COVID           & 84.29 &	81.24 &	83.22 & 82.92 (1.27) & 80.39 &	79.03 &	80.69 & 80.04 (1.38) \\
SciFact + Check-COVID         & 83.51 &	79.49 &	80.64 & 81.21 (1.06) & 80.03 &	82.95 &	83.49 & 82.16 (1.14)\\
FEVER + SciFact + Check-COVID & 84.17 &	81.13 &	84.66 & \textbf{83.32}\dag (2.01) & 78.94 &	76.89 &	81.79 & 79.21 (2.64) \\
COVID-Fact + Check-COVID     & 83.75 &	77.75 &	82.86 & 81.46 (0.99)& 72.69 &	82.49 &	81.56 & 78.91 (2.97)\\
\midrule
GPT-3.5 (text-davinci-003) & 60.22 & 70.71 & 75.00 & 68.64 & 33.33 & 54.55 & 66.67 & 51.52\\

\bottomrule
\end{tabular}
\end{adjustbox}
\caption{F1 (avg. of $3$ seeds, std in parentheses) of our \textit{Label Prediction} module under different training configurations (combinations of Check-COVID, FEVER, SciFact and COVID-Fact; \textit{composed} and \textit{extracted} claims alone and together). } \label{tabel_covid-lab}
\end{table*}

\begin{table}[]
\centering
\begin{adjustbox}{max width=0.42\textwidth}
 \begin{tabular}{cc|ccc}
\toprule

 \multicolumn{2}{l|}{\textbf{\makecell{Single Sentence\\Rationale Investigation}  }}          & \makecell{Predicted\\Rationale}  & \makecell{Sampled Gold\\Rationale}\\
\midrule
\multirow{2}{*}{\textit{Allow NEI}}  & Accuracy & 85.71              & 50.00         \\
                                    & Macro F1 & 58.04              & 44.85         \\
\multirow{2}{*}{\textit{Ignore NEI}} & Accuracy & 92.86              & 85.71         \\
                                    & Macro F1 & 92.51              & 85.71 \\
\bottomrule
\end{tabular}
\end{adjustbox}
\caption{When a dev set gold rationale contains many sentences but our model only selects one, we compare \textit{Label Prediction} performance using the selected sentence vs. a randomly sampled gold rationale sentence.} \label{tabel_ablation}
\end{table}

\section{Experiments}
\label{sec_experiments}
To establish a baseline score for Check-COVID, we first evaluate the dev-set performance of our \textit{Rationale Selection} and \textit{Label Prediction} modules. We then choose the best-performing model for each to build the full end-to-end pipeline system which we evaluate on the test split. Given the material differences between \textit{composed} and \textit{extracted} claims, we evaluate training on each claim type separately and in tandem. As Vespa could return many relevant abstracts from CORD-19 beyond the ones selected for our corpus, we omit evaluation of the \textit{Vespa}-based \textit{Abstract Retrieval} module by itself. 

\textbf{Training Datasets} To build our \textit{Rationale Selection} and \textit{Label Prediction} modules, we fine-tune RoBERTa-Large models sequentially on FEVER and/or SciFact and then on Check-COVID. We note that though FEVER and SciFact contain claims and evidence unrelated to COVID-19, their inclusion allows us to evaluate the transfer potential of additional task-related data from different domains.  For \textit{Label Prediction}, we also explore fine-tuning on COVID-Fact (which we cannot use for \textit{Rationale Selection} as it does not contain full abstracts). As we only use these corpora for research, it is consistent with their intended use.

\textbf{Training Setting} While training the \textit{Rationale Selection} module, we pair each claim with its labeled evidence sentences to produce positive examples from FEVER, SciFact, or Check-COVID;
 we create negative examples by pairing claims and non-evidence sentences from the same abstracts that contained the labeled evidence. Since \texttt{NOTENOUGHINFO} claims are not paired with relevant evidence, we only use \texttt{SUPPORT}/\texttt{REFUTE} data points for training this module.  For the \textit{Label Prediction} module, we train on each claim and the concatenation of its gold-label evidence sentences. 

\textbf{Hyperparameter Settings} For the rationale and label prediction modules, we use batch sizes of $256$ and $16$ respectively.  
We use   learning rates of 1e-5 for our RoBERTa encoders and 1e-4 for our linear classifier heads. We train on $4$ V100 GPUs up to $20$ epochs with early stopping (within $6$ epochs).\footnote{Results are averaged across three random seeds} 

 \begin{table*}[]
\centering
\begin{adjustbox}{max width=0.8\textwidth}
\begin{tabular}{llccc|ccc}
\toprule
&     & \multicolumn{3}{c}{\textbf{Oracle}}  & \multicolumn{3}{c}{\textbf{Vespa}}  \\
Train / Dev & Test  &  \texttt{REFUTE}  & \texttt{SUPPORT} & Macro F1 & \texttt{REFUTE}  & \texttt{SUPPORT} & Macro F1 \\
\midrule
 \multicolumn{6}{l}{\textit{* Allow \texttt{NOTENOUGHINFO} predictions}}  \\
Composed    & \multirow{2}{*}{Composed}  & 87.80 &	86.27 & 87.04 & 58.59 &	68.09 & \textbf{63.34} \\
Comp + Extr &                            & 83.76 &	86.27 & 85.02 & 54.00 &	62.22 & 58.11 \\
\midrule
Extracted   & \multirow{2}{*}{Extracted} & 95.65 &	88.89 & \textbf{92.27} & 40.00 &	59.26 & 49.63\\
Comp + Extr &                            & 85.71 &	87.18 & 86.45 & 28.57 &	59.26 & 43.92 \\
\midrule
\midrule
 \multicolumn{6}{l}{\textit{* Ignore \texttt{NOTENOUGHINFO} predictions}}  \\
Composed    & \multirow{2}{*}{Composed}  & 92.19 &	90.91 & \textbf{91.55} & 79.70 &	74.29 & \textbf{76.99} \\
Comp + Extr &                            & 90.91 & 88.68 & 89.79 & 74.42 &	69.72 & 72.07 \\
\midrule
Extracted   & \multirow{2}{*}{Extracted} & 88.00 &	91.89 & 89.95 & 56.00 &	70.27 & 63.14 \\
Comp + Extr &                            & 85.71 &	92.68 & 89.20 &25.00 &	73.91 & 49.46 \\
\midrule
  \multirow{2}{*}{GPT-3.5 (text-davinci-003)}  & Composed & 63.92 & 75.18 & 69.55 & 74.45 & 65.35 & 69.90\\
 & Extracted & 40.00  &  80.85 & 60.43 & 62.5 & 60.00 & 61.25\\

\bottomrule
\end{tabular}
\end{adjustbox}
\caption{Full fact-checking pipeline F1 with different configurations of \textit{composed} and \textit{extracted} claims. \textit{Oracle}/\textit{Vespa} indicates abstract retrieval method. \textit{Rationale Selection} and \textit{Label Prediction} uses Tables \ref{tabel:covid-rat} and \ref{tabel_covid-lab} \dag-ed models. 
}
\label{tabel_covid-pipeline}
\end{table*}
\vspace{-3mm}

\subsection{Rationale Selection Evaluation}
\label{sec:rateval}
We evaluate predicted rationales on the Check-COVID dev set by considering rationale sentences individually and in aggregate. When considering a rationale's sentences individually (sentence level), we calculate two scores: \textbf{standard} precision and recall and the \textbf{global recall-focused} (GRF) variant of precision and recall used by the SciFact paper. GRF scores only consider a selected sentence for a claim to be correct if all of the claim's gold sentences are also selected. While we include this score, we show that our full pipeline performs equally well even in cases where all gold sentences are not predicted by the \textit{Rationale Selection} module, suggesting that this focus on recall may be misplaced.

When considering each rationale in aggregate (rationale level), we calculate two different scores: a \textbf{strict} score where a predicted rationale is correct only if it is identical to the gold (no additional/missing sentences) and
a more permissive \textbf{intersection} score where a predicted rationale is correct if it contains at least one gold sentence. As we shall see, predicted rationales that only intersect with the gold rationale still result in good downstream performance on label prediction.

\textbf{Results} In Table \ref{tabel:covid-rat}, we present the baseline performance for the \textit{Rationale Selection} module on Check-COVID's dev set. When evaluating on \textit{composed} claims, models trained on both \textit{composed} and \textit{extracted} claims outperformed models trained on \textit{composed} alone.  The model fine-tuned on only Check-COVID and the model fine-tuned on FEVER and Check-COVID performed best on the harder evaluation metrics: $38.31$ for \textit{global recall-focused} F1 at the sentence level and $31.13$ for \textit{strict} accuracy at the rationale level respectively.  However, on the easier metrics, the model fine-tuned on FEVER, SciFact and Check-COVID together scored highest: $52.18$ for \textit{standard} F1 at the sentence level and $75.16$ for \textit{intersection} accuracy at the rationale level.    

Interestingly, we observe that including \textit{composed} claims during fine-tuning degrades performance on \textit{extracted}. Training on FEVER and Check-COVID produces the best scores on \textit{extracted} across both easier and harder metrics.

We note that for both \textit{composed} and \textit{extracted} claims, our best performing models exhibit a $40$-point improvement when evaluating with \textit{intersection} accuracy instead of \textit{strict}. It is possible that these high \textit{intersection} scores are masking cases where the \textit{Rationale Selection} module is picking less informative evidence. To understand this better, in Table \ref{tabel_ablation} we show that even when the model only chooses one sentence from a multi-sentence rationale, it is picking high quality evidence. Scores with the model's selections outperform randomly sampled gold evidence by a wide margin.

Finally, our results show that fine-tuning on FEVER in addition to Check-COVID most often boosts performance while SciFact does not always help. This suggests that while the large size of FEVER helps, the genre difference between SciFact claims (from journals) and Check-COVID claims (from news) limits effective transfer.

\subsection{Label Prediction Evaluation}

We evaluate the \textit{Label Prediction} module on the Check-COVID dev set using standard macro-F1.

\textbf{Results} The baseline results for the \textit{Label Prediction} module are presented in Table \ref{tabel_covid-lab}. For \textit{composed} claims, we note that the model trained on FEVER + SciFact + Check-COVID (\textit{composed} training examples) achieves the best performance ($83.76$ macro F1). 
For \textit{extracted} claims, training on FEVER + Check-COVID (\textit{extracted} examples) achieves the highest score ($85.19$ macro F1). We suspect that FEVER exhibits the best transfer performance due to its large size and its inclusion of the \texttt{NOTENOUGHINFO} veracity class, however, there is still considerable room for improvement, perhaps because the evidence in FEVER is drawn from Wikipedia rather than scientific journals. Additionally, training on both \textit{composed} and \textit{extracted} examples does not improve performance in most settings when compared to training on \textit{composed} or \textit{extracted} claims alone.

In Table \ref{tabel_covid-lab}, we present the models' performance broken out by label.
We note that performance on \texttt{REFUTE} claims is similar to performance on \texttt{SUPPORT} claims on average, demonstrating that \texttt{REFUTE} claims are not easier than \texttt{SUPPORT} claims for our models. As \texttt{REFUTE} claims are manually negated by our annotators, this suggests this process does not introduce spurious signals that our models learn to exploit.

\subsection{Pipeline Evaluation}
We evaluate the end-to-end performance of the full fact-checking pipeline on each of the Check-COVID \textit{composed} and \textit{extracted} test sets.  We do so under two settings: using \textit{Oracle} (i.e., gold) abstracts and using \textit{Vespa} to retrieve abstracts. We use the \textit{Rationale Selection} and \textit{Label Prediction} models that performed best on the Check-COVID dev set (see the $\dag$-ed numbers in Tables \ref{tabel:covid-rat} and \ref{tabel_covid-lab}). For this evaluation setting we do not allow the \textit{Rationale Selection} model to produce empty rationales.

Our labeled \texttt{NOTENOUGHINFO} examples require passing insufficient evidence to the \textit{Label Prediction} module. However, the \textit{Oracle} abstracts always contain sufficient evidence. Additionally, it is difficult to know \textit{a priori} whether \textit{Vespa} is returning abstracts that lack sufficient evidence. This is primarily because, besides the ones we've selected, the CORD-19 dataset likely includes many more abstracts that contain evidence relevant to any given claim.
Therefore, when evaluating the full pipeline, we remove examples labeled \texttt{NOTENOUGHINFO} from our test data. Because our label prediction model was trained on all $3$ veracity labels, we evaluate it under two conditions: 1) we allow the model to predict \texttt{NOTENOUGHINFO} (\textit{allow NEI}), and 2) we select the output class by only considering the logits for the \texttt{SUPPORTS} and \texttt{REFUTES} indices in the predictor's output (\textit{ignore NEI}). We use the 2-class (\texttt{SUPPORT}/\texttt{REFUTE}) macro F1 scores of the \textit{Label Prediction} module as the final score for each full pipeline variant.

\textbf{Results} We present the full pipeline results in Table \ref{tabel_covid-pipeline}. First, consider the \textit{Oracle} abstract setting. For \textit{composed} claims, the best performance ($87.04$ macro F1 for \textit{allow NEI}, $91.55$ for \textit{ignore NEI}) is achieved by training on only \textit{composed} examples. Likewise, for \textit{extracted} claims, training on \textit{extracted} claims produces the best performance ($92.27$ macro F1 for \textit{allow NEI}, $89.95$ for \textit{ignore NEI}). These results show that datasets designed for one type of claim do not necessarily transfer perfectly to the other, even within the same domain.  
When considering the performance of \textit{Vespa} as our \textit{Abstract Retrieval} module, we observe a fairly large drop in F1 ($20$ - $30$ points in most settings).  Additionally, the \textit{Vespa}-based full pipelines benefit more from ignoring when the \textit{Label Prediction} module predicts \texttt{NOTENOUGHINFO}. This suggests that \textit{Vespa} may be retrieving less relevant abstracts that result in more predictions of \texttt{NOTENOUGHINFO}. This result demonstrates the importance of including a \texttt{NOTENOUGHINFO} label, as an \textit{Oracle} for retrieving relevant abstracts does not exist in the wild.

Finally, the models for \textit{Rationale Selection} and \textit{Label Prediction} achieve similar performance on both \textit{composed} and \textit{extracted} claims despite the complexity of the \textit{extracted} claims' surface forms. This suggests that the \textit{Vespa} pipelines' degraded performance on \textit{extracted} claims was due to \textit{Vespa} struggling to retrieve evidence for the difficult news claims in the \textit{extracted} subset.

\subsection{In-Context Learning with GPT-3.5}

Due to the success of in-context learning across a wide range of NLP benchmarks, we additionally evaluate GPT-3.5 \cite{brown2020language} on Check-COVID using a few-shot setting.\footnote{Experimental details can be found in the Appendix \ref{sec:prompt}.}  As is evident from Tables \ref{tabel:covid-rat}, \ref{tabel_covid-lab} and \ref{tabel_covid-pipeline}, GPT-3.5 performs worse than our trained baseline on rationale selection (by $10$-$20$ points), label prediction (by $15$-$30$ points) and in the full pipeline setting (by $6$-$20$ points). It is possible that without task-specific fine-tuning, it struggles to ground the natural language claims in scientific jargon.

\section{Conclusion}

We present Check-COVID, a new corpus for fact-checking everyday claims about COVID-19 against evidence from scientific journal articles.  While our experimental results establish a strong baseline, they also demonstrate the difficulty in transferring learning from existing fact-checking corpora to this new dataset. In future work, we plan to train a unified model to perform rationale selection and label prediction, mitigating error propagation.

\section*{Limitations}

Due to time and budget constraints, this work remains limited in a number of important ways.  The relatively small size of our corpus and its specificity to COVID-19 necessitates the development of systems with richer inductive biases and the ability to effectively transfer knowledge from related corpora like FEVER.  Additionally, due to the large size of CORD-19, the database of scientific literature from which we draw the abstracts in Check-COVID, it is difficult to evaluate abstract retrieval components like Vespa with our annotations.  There are likely many abstracts in CORD-19 in addition to the ones we've selected that contain evidence relevant to any given claim, precluding both measurements of abstract precision and the evaluation of \texttt{NOTENOUGHINFO} in the full fact-checking pipeline setting. Finally, as the claims in Check-COVID are drawn from western, English language news sources and annotators, they are likely unrepresentative of the full range of COVID-related (mis)information in need of fact-checking online. 
\section*{Ethics Statement}
There are a few ethical considerations to consider if someone were to use our fact-checking system in the real world. Fact-checking results are entirely dependent on the source of truth used. In this work, we use peer-reviewed scientific journal articles, but a malicious actor could easily swap the backend of our system (Vespa) and instead have it search over some source of information that advances that actor's agenda. This type of use would further propaganda and misinformation. Additionally, many fact-checking processes are used to flag and potentially remove content online. Content moderation has many positive effects in promoting healthy online communities, but again, in the wrong hands with the wrong source of truth, this type of system could be used for unjust censorship. Finally, our claims were collected from news sites that lean toward the left or middle on the United States political spectrum\footnote{\url{https://www.allsides.com/media-bias/media-bias-chart}}. The system could therefore behave differently on claims from more right-leaning news sources in a way that might favor one set of views over another. 
\section*{Acknowledgements}
This research was developed with funding from the Defense Advanced Research Projects Agency (DARPA) under Contract No. HR001120C0123. The views, opinions and/or findings expressed are those of the author and should not be interpreted as representing the official views or policies of the Department of Defense or the U.S. Government.

\bibliography{anthology,custom}
\bibliographystyle{acl_natbib}

\appendix
\section{News Citance Sources}
\label{sec:News_Citance_Sources}
We extracted citances that contain URLs to medical journals from following news websites: The New York Times, The Washington Post, The Atlantic, CNN, NPR, BBC. We note that while these websites are trustworthy and broadly relied upon by fact-checkers, most of them exhibit a bias to the political left which could constrain the breadth of claims we collect. 

\section{Annotation Interface \\ and Instruction Guide}
\label{sec:annotation_interface}
\begin{figure*}[h] 
\includegraphics[scale=0.2]{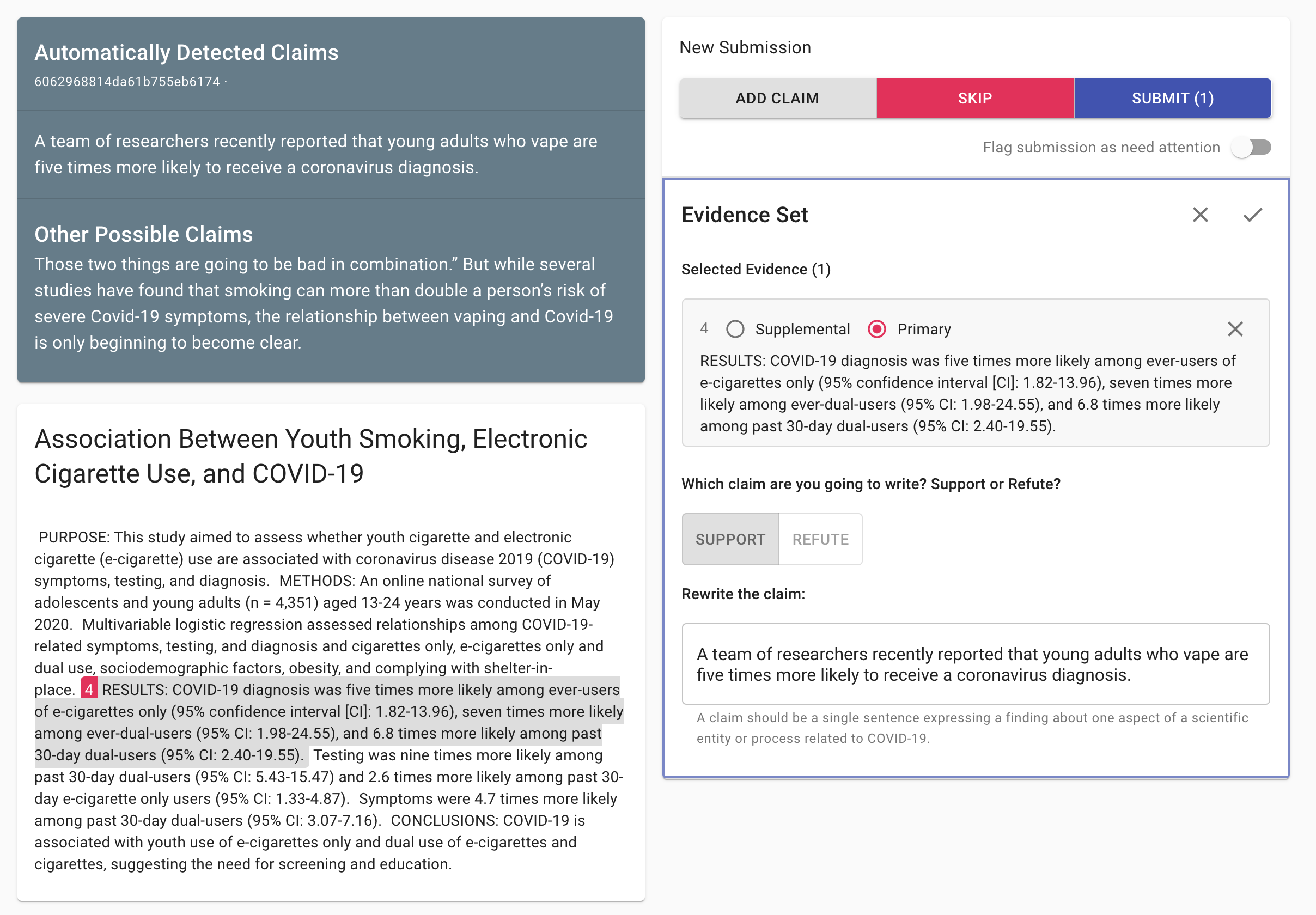}
\centering
\caption{Annotation Interface}
\label{figure:annotation_interface}
\end{figure*}

We present the annotation interface in Figure \ref{figure:annotation_interface}. The detailed annotation instruction guide is available at

\url{https://drive.google.com/file/d/1q6diZgcJquxBZMHdViYn34d8DMVEMVV9/view}.
The annotation instruction video is available at \url{https://youtu.be/mEOcouML9oA}.

\section{Experiment with GPT-3.5} 
\label{sec:prompt}
In our studies involving GPT-3.5, we devised prompts for the module and pipeline evaluation. These prompts comprised a random selection of 2 or 3 examples for in-context learning. For the sake of brevity in this paper, we substituted the actual examples in the prompt templates with the placeholder "example abstract/claim/rational sentence". Our work was carried out on the text-davinci-003 variant of GPT-3.5, using the completion API endpoint with hyperparameters, including a temperature setting of 0.7, a cap on the token count at 256, a top-p value fixed at 1, and both frequency and presence penalties set to 0.

\textbf{Prompt for Rationale Selection in Module Evaluation}\\

List id(s) of sentence(s) in the abstract that support or refute the claim if they exist. If none apply, return empty list.\\
\\
Abstract:\\
0: example abstract sentence\\
... example abstract sentence \\
7: example abstract sentence\\
Claim: example claim sentence\\
Selected id(s): [7]\\
\\
Abstract:\\
0: example abstract sentence\\
... example abstract sentence\\
6: example abstract sentence\\
Claim: example claim sentence\\
Selected id(s): [4, 5]\\
\\
Abstract: \\
\{abstract\}\\
Claim: \{claim\}\\
Selected id(s):\\
\\
\textbf{Prompt for Rationale Selection in Pipeline Evaluation}\\
List id(s) of sentence(s) in the abstract that support or refute the claim if they exist. If none apply, return at least one that is most related to the claim.\\
\\
Abstract:\\
0: example abstract sentence\\
... example abstract sentence \\
7: example abstract sentence\\
Claim: example claim sentence\\
Selected id(s): [7]\\
\\
Abstract:\\
0: example abstract sentence\\
... example abstract sentence\\
6: example abstract sentence\\
Claim: example claim sentence\\
Selected id(s): [4, 5]\\
\\
Abstract: \\
\{abstract\}\\
Claim: \{claim\}\\
Selected id(s):\\
\\
\textbf{Prompt for Label Prediction in Module Evaluation}\\
Decide whether the rationale refutes or supports the claim or if there is not enough info to make a decision on the claim.\\
\\
Claim: example supported claim sentence\\
Rationale: example rationale sentences that support the claim\\
Label: SUPPORT\\
\\
Claim: example refuted claim sentence\\
Rationale: example rationale sentences that refute the claim\\
Label:  REFUTE\\
\\
Claim: example claim sentence\\
Rationale: example rationale sentences that do not include enough information about the claim to support or refute\\
Label:  NOT ENOUGH INFO\\
\\
Claim: \{claim\}\\
Rationale: \{evidence\}\\
Label: \\
\\
\textbf{Prompt for Label Prediction in Pipeline Evaluation}\\
Decide whether the rationale refutes or supports the claim.\\
\\
Claim: example supported claim sentence\\
Rationale: example rationale sentences that support the claim\\
Label: SUPPORT\\
\\
Claim: example refuted  claim sentence\\
Rationale: example rationale sentences that refute the claim\\
Label: REFUTE\\
\\
Claim: \{claim\}\\
Rationale: \{evidence\}\\
Label:

\end{document}